\documentclass[10pt,twocolumn,letterpaper]{article}

\usepackage{iccv}
\usepackage{times}
\usepackage{epsfig}
\usepackage{graphicx}
\usepackage{amsmath}
\usepackage{amssymb}

\usepackage{epstopdf}

\usepackage[breaklinks=true,bookmarks=false]{hyperref}

\usepackage{multicol}
\usepackage{multirow}

\newcommand{\Section}[1]{\vspace{-2pt}\section{#1}\vspace{-2pt}} 
\newcommand{\Paragraph}[1]{\vspace{-4pt}\paragraph{#1}} 

\newcommand*{\vv}[1]{\vec{\mkern0mu#1}}

\iccvfinalcopy 

\def\iccvPaperID{****} 

\ificcvfinal\fi

\iccvfinalcopy 

\def\iccvPaperID{****} 

\ificcvfinal\fi

\begin{document}

\title{ HEMlets Pose: Learning Part-Centric Heatmap Triplets \\ for Accurate 3D Human Pose Estimation}

\author{Kun Zhou$^1$,~~Xiaoguang Han$^2$,~~Nianjuan Jiang$^1$,~~Kui Jia$^3$,~~and Jiangbo Lu$^{1*}$\\
$^1${Shenzhen Cloudream Technology Co., Ltd.}~~~$^2${The Chinese University of Hong Kong (Shenzhen)}\\
$^3${South China University of Technology}~~~$^*${Corresponding email: {\small jiangbo.lu@gmail.com}}\\
}


\maketitle
\ificcvfinal\thispagestyle{empty}\fi

\begin{abstract}
   Estimating 3D human pose from a single image is a challenging task. This work attempts to 
    address the uncertainty of lifting the detected 2D joints to the 3D space by introducing an 
    intermediate state - Part-Centric Heatmap Triplets ({\emph {HEMlets}}), which shortens 
    the gap between the 2D observation and the 3D interpretation. The HEMlets utilize 
    three joint-heatmaps to represent the relative depth information of the end-joints 
    for each skeletal body part. In our approach, a Convolutional Network~(ConvNet) is first trained to predict 
    HEMlets from the input image, followed by a volumetric joint-heatmap regression. 
    We leverage on the integral operation to extract the joint locations from the volumetric 
    heatmaps, guaranteeing end-to-end learning. Despite the simplicity of the network design, the 
    quantitative comparisons show a significant performance improvement over the 
    best-of-grade method (about $20\%$ on Human3.6M). The proposed method naturally supports training 
    with ``in-the-wild'' images, where only weakly-annotated relative depth information of skeletal 
    joints is available. This further improves the generalization ability of our model, as 
    validated by qualitative comparisons on outdoor images.
\end{abstract}

\Section{Introduction}

Human pose estimation from a single image is an important problem in computer vision, 
because of its wide applications, e.g., video surveillance and human-computer interaction. 
Given an image containing a single person, 3D human pose inference aims to predict 
3D coordinates of the human body joints. Recovering 3D information of human poses 
from a single image faces several challenges. The challenges are at least three 
folds: 1) reasoning 3D human poses from a single image is by itself 
very challenging due to the inherent ambiguities; 
2) being a regression problem, existing approaches have not achieved 
a good balance between representation efficiency and learning effectiveness;
3) for ``in-the-wild" images, both 3D capturing and manual labeling require 
a lot of efforts to obtain high-quality 3D annotations, making the 
training data extremely scarce. 

For 2D human pose estimation, almost all best performing methods are detection 
based~\cite{SH-NET,ke2018multi,wei2016convolutional}. Detection-based approaches essentially 
divide the joint localization task into local image classification tasks. The latter 
is easier to train, because it effectively reduces the feature and target dimensions 
for the  learning system~\cite{sun2017integral}. 
Existing 3D pose estimation methods often use detection as an intermediate 
supervision mechanism as well. A straightforward strategy is to 
use volumetric heatmaps to represent the likelihood map of each 3D 
joint location~~\cite{pavlakos2017coarse}. Sun~{\it et al}.~\cite{sun2017integral} 
further proposed a differentiable soft-argmax operator that unifies the joint detection 
task and the regression task into an end-to-end training framework. 
This significantly improves the state-of-the-art 3D pose estimation accuracy. 

\begin{figure}
  \centering
  \includegraphics[width=\columnwidth]{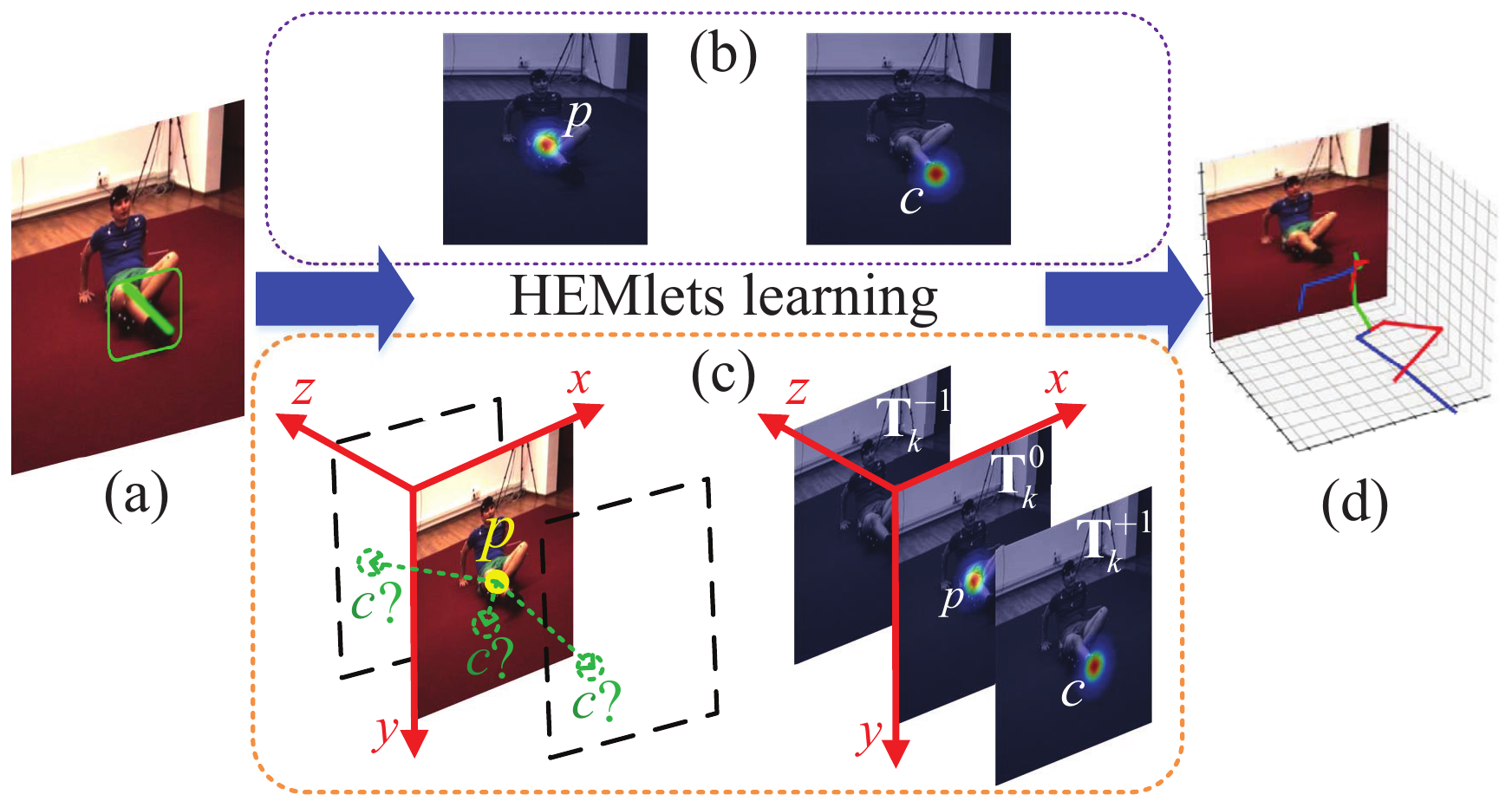}
	\caption{Overview of the HEMlets-based 3D pose estimation. (a)~Input RGB image. Our algorithm encodes (b)~the 2D locations for the joints $p$ and $c$, but also (c)~their relative depth relationship for each skeletal part $\vv{pc}$ into HEMlets. (d) Output 3D human pose.}
  \label{fig:teaser}
\end{figure}

In this work, we propose a novel effective intermediate representation for 3D pose estimation - 
\emph{Part-Centric Heatmap Triplets (HEMlets)} (as shown in Fig.~\ref{fig:teaser}). 
The key idea is to polarize the 3D volumetric space around each 
distinct skeletal part, which has the two end-joints kinematically connected. 
Different from 
\cite{pavlakos2018ordinal}, our relative depth information is represented as three polarized heatmaps, 
corresponding to the different state of the local depth ordering of the part-centric joint pairs. Intuitively, HEMlets encodes the co-location 
likelihoods of pairwise joints in a dense per-pixel manner with the coarsest 
discretization in the depth dimension. Instead of considering arbitrary joint pairs, we focus on kinematically connected 
ones as they possess semantic correspondence with the input image, and are thus a more effective target for the subsequent learning. 
In addition, the encoded relative depth information is strictly local for the part-centric joint pairs and suffers less from potential inconsistent data annotations. 

The proposed network architecture is shown in Fig.~\ref{fig:framework}. A ConvNet is first 
trained to learn the HEMlets and 2D joint heatmaps, which are then fed together with the high-level image features to another ConvNet to produce a volumetric 
heatmap for each joint. We leverage on the soft-argmax regression~\cite{sun2017integral} 
to obtain the final 3D coordinates of each joint. Significant improvements 
are achieved compared to the best competing methods quantitatively and qualitatively. 
Most notably, our HEMlets method achieves a record MPJPE of 39.9mm on Human3.6M~\cite{ionescu2014human3}, yielding about $20\%$ improvement over the best-of-grade method~\cite{sun2017integral}. 

The merits of the proposed method lie in three aspects:
\begin{itemize}
  \vspace{-4pt}
  \item 
  \textbf{\textit{Learning strategy. }}
  Our method takes on a progressive learning strategy, and 
  decomposes a challenging 3D learning task into a sequence of easier sub-tasks with mixed 
  intermediate supervisions, i.e., 2D joint detection and HEMlets learning. HEMlets 
  is the key bridging and learnable component leading to 3D heatmaps, and is much easier to 
  train and less prone to over-fitting. Its training can also take advantage of existing labeled 
  datasets of relative depth ordering~\cite{pavlakos2018ordinal,shi2018fbi}. 
	\vspace{-4pt}
  \item
  \textbf{\textit{Representation power.}}~HEMlets is based on 2D per-joint heatmaps, but extends them by a couple of additional heatmaps 
  to encode local depth ordering in a dense per-pixel manner. It builds on top of 2D heatmaps 
  but unleashes the representation power, while still allowing leveraging the ssoft-argmax 
  regression~\cite{sun2017integral} for end-to-end learning. 
	\vspace{-4pt}
  \item 
  \textbf{\textit{Simple yet effective. }}
  The proposed method features a simple network architecture design, and it is easy to train and 
  implement. It achieves state-of-the-art 3D pose estimation results validated by the evaluations over all standard benchmarks. 
\end{itemize}

\Section{Related Work}\label{sec:relatedWork}
In this section, we review the approaches that are based on deep ConvNets for 3D human pose estimation.

\subsection{Direct encoder-decoder}
With the powerful feature extraction capability of deep ConvNets, many approaches~\cite{li20143d,tekin2016structured,park20163d} learn end-to-end 
\textit{Convolutional Neural Networks}~(CNNs) to infer human poses directly from the images. Li and Chen~\cite{li20143d} are 
the first who used CNNs to estimate 3D human pose via a multi-task framework. Tekin~{\it et al}.~\cite{tekin2016structured} designed an auto encoder to model the joint dependencies in a high-dimensional feature space. Park~{\it et al}.~\cite{park20163d} proposed fusing 2D joint locations with high-level image features to boost the estimation of 3D human pose. However, these single-stage 
methods are limited by the availability of 3D human pose datasets and cannot take advantage of large-scale 2D pose datasets that are vastly available.

\subsection{Transition with 2D joints}

To avoid collecting 2D-3D paired data, a large number of works~\cite{ronchi2018s,zhou2017towards,yang20183d,martinez2017simple,fang2018learning,shi2018fbi} decompose the task of 3D pose estimation into two independent stages: 1) firstly inferring 2D joint locations using well-studied 2D pose estimation methods, such as~\cite{zhou2017towards,ronchi2018s}; 2) and then learning a mapping to lift them into the 3D space. These approaches mainly focus on tackling
the second problem. For example, a simple fully connected residual network is proposed by Martinez~{\it et al}.~\cite{martinez2017simple} to directly recover 3D human pose from its 2D projection. Fang~{\it et al}.~\cite{fang2018learning} considered prior knowledge of human body configurations and proposed human pose grammar, leading to better recovery of the 3D pose from only 2D joint locations. Yang~{\it et al}.~\cite{yang20183d} adopted an adversarial learning scheme to ensure the anthropometrical validity of the output pose and further improved the performance. Recently, by involving a reprojection mechanism, the proposed method in~\cite{wandt2019repnet} shows insensitivity to overfitting and 
accurately predicts the result from noisy 2D poses. Though promising results have been achieved by these two-stage methods, 
a large gap exists between the 3D human pose and its 2D projections due to inherent ambiguities.

\subsection{3D-aware intermediate states}

To further bridge the gap between the 2D image and the target 3D human pose under estimation, 
some recent works~\cite{pavlakos2017coarse,shi2018fbi,pavlakos2018ordinal,sun2017integral} proposed to involve 3D-aware states for intermediate supervisions. Namely, a network is firstly trained to map the input image to these 3D-aware states, and then another network is trained to convert those states to the 3D joint locations. Finally, these two networks are combined and optimized jointly. A volumetric representation for 3D joint-heatmaps is proposed in ~\cite{pavlakos2017coarse}, with which the 3D pose is regressed in a coarse-to-fine manner. However, regressing a probability grid in the 3D space globally is also a very challenging task. It usually suffers from quantization errors for the joint locations. To address this issue, Sun~{\it et al}.~\cite{sun2017integral} exploited a soft-argmax operation and proposed 
an end-to-end training scheme for the 3D volumetric regression, 
achieving by far the best performance on 3D pose estimation. 
Inspired by~\cite{pons2014posebits} that the relative depth ordering 
across joints is helpful for resolving pose ambiguities, Pavlakos~{\it et al}.~\cite{pavlakos2018ordinal} adopted a ranking loss for pairwise ordinal depth to train the 3D human pose predictor explicitly. A similar scheme of relative depth supervision is utilized in the work of~\cite{ronchi2018s}. Forward-or-Backward Information~(FBI), proposed in ~\cite{shi2018fbi}, is another kind of relative depth information 
but focuses more on the bone orientations. 

In this work, we propose HEMlets, a novel representation that encodes both 
2D joint locations and the part-centric relative depth ordering simultaneously. Experiments justify that this representation reaches by far the best balance between representation efficiency and learning effectiveness. 

\begin{figure}[t]
\centering 
\includegraphics[width=\linewidth]{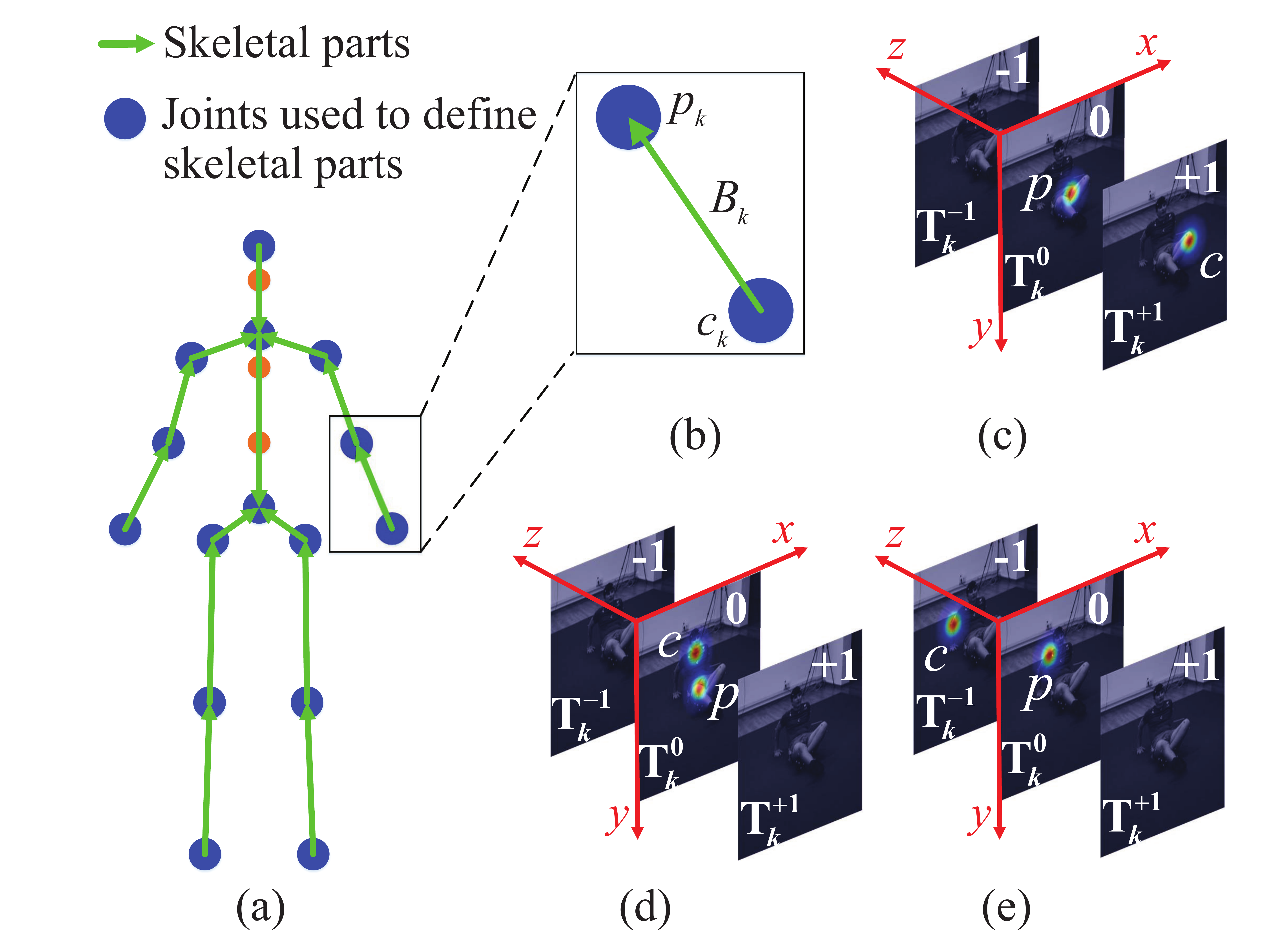}
\caption{Part-centric heatmap triplets $\{{\bf T}^{-1}_k,{\bf T}^{0}_k,{\bf T}^{+1}_k\}$ where $p$ and $c$ are the parent joint and the child joint. (a,~b) Joints and skeletal parts. We locate the parent joint $p$ of the $k$-th skeletal part $B_k$ at the zero polarity heatmap ${\bf T}^{0}_k$~(c-e). The child joint $c$ is located, according to relative depth of $p$ and $c$, in the positive~(c), zero~(d) and negative polarity heatmap~(e), respectively. }
\label{fig:tri_heatmaps}
\end{figure}

\subsection{``In-the-wild'' adaptation}
All the aforementioned approaches are mainly trained on the datasets collected under indoor settings, due to the difficulty of annotating 3D joints for ``in-the-wild'' images ~\cite{bourdev2009poselets}. 
Thus, many strategies are developed to make domain adaptation. 
By exploiting graphics techniques, previous works~\cite{varol2017learning,chen2016synthesizing} have synthesized a large ``faked'' dataset mimicking real images. Though these data benefit 3D pose estimation, they are still far from realistic, making the applicability limited. Recently, both Pavlakos~{\it et al}.~\cite{pavlakos2018ordinal} and Shi~{\it et al}.~\cite{shi2018fbi}  proposed to label the relative depth relationship across joints 
instead of the exact 3D joint coordinates. This weak annotation scheme 
not only makes building large-scale ``in-the-wild'' datasets 
feasible, but also provides 3D-aware information 
for training the inference model in a weakly-supervised manner. 
With HEMlets representation, we can readily use these weakly annotated ``in-the-wild'' data for domain adaptation.

\begin{figure*}[t]
\centering
\vspace{-0.15in}
\includegraphics[width=16.5cm]{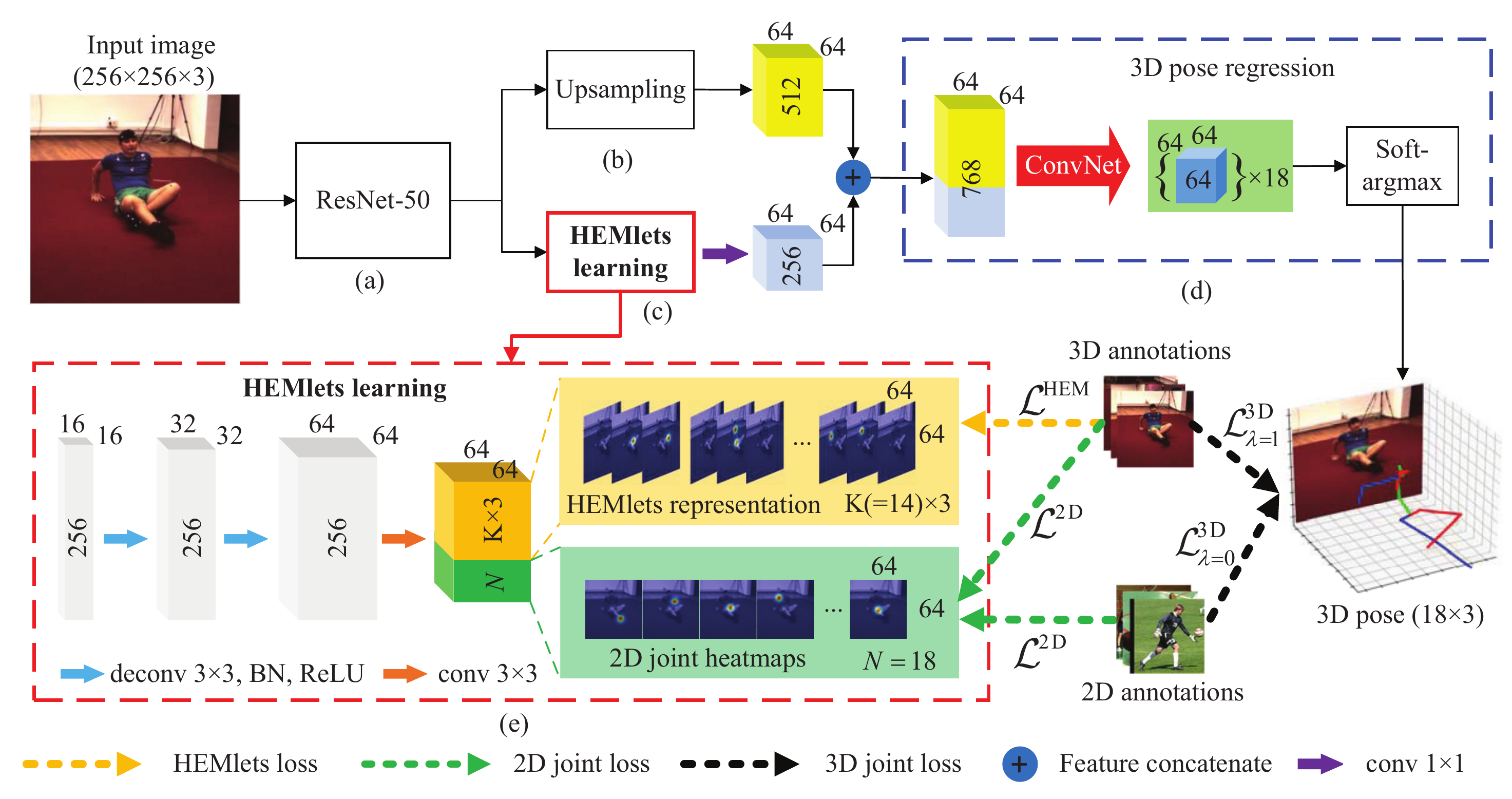}
\caption{The network architecture of our proposed approach. It consists of 
four major modules: (a) A ResNet-50 backbone for image feature extraction. (b) A ConvNet for image feature upsampling. (c) Another ConvNet for HEMlets learning and 2D joint detection. 
(d) A 3D pose regression module adopting a soft-argmax operation for 3D human pose estimation. (e) Details of the HEMlets learning module. ``Feature concatenate" denotes concatenating the feature maps from the HEMlets learning branch and the upsampling branch together.}
\label{fig:framework}
\end{figure*}

\Section{HEMlets Pose Estimation} \label{sec:method}

We propose a unified representation of heatmap triplets to model the local information of body skeletal parts, i.e., kinematically connected joints, whereas the corresponding 2D image coordinates and relative depth ordering are considered. By such a representation, images annotated with relative depth ordering of skeletal parts can be treated equally with images annotated with 3D joint information. While the latter is usually very scarce, the former is relatively easy to obtain~\cite{shi2018fbi,pavlakos2018ordinal}. In this section, we first present the proposed part-centric heatmap triplets and its encoding scheme. Then, we elaborate a simple network architecture that utilizes the part-centric heatmap triplets for 3D human pose estimation. \\


\subsection{Part-centric heatmap triplets}
\label{sec:constructHemlets}

We divide the full body skeleton consisting of $N=18$ joints into $K=14$ parts as shown in Fig.~\ref{fig:tri_heatmaps}(a).
Specifically, we use $B$ to denote the set of skeletal parts,
where $B = \left\{ B_1,B_2,\ldots,B_K \right\}$. For each part,
we denote the two associated joints as $(p, c)$, with $p$ being the parent node and $c$ being the child node. The relative depth ordering, denoted as $r(z_p, z_c)$, can be then described as a tri-state function~\cite{pavlakos2018ordinal,shi2018fbi}:
\begin{equation} 
r(z_p,z_c) =
\begin{cases}
1& \text{$z_p - z_c>\epsilon$}\\
0& \text{$\left| z_p - z_c \right|<\epsilon$} \\
-1& \text{$z_p - z_c<-\epsilon$}
\end{cases} ,
\label{eq:polarity}
\end{equation}

\noindent where $\epsilon$ is used to adjust the sensitivity of the function to the relative depth difference. The absolute depths of the two joints $p$ and $c$  are denoted by $z_p$ and $z_c$, respectively.

We argue that directly using the discretized label as an intermediate state for learning the
3D pose from a 2D joint heatmap, as was done in \cite{pavlakos2018ordinal,shi2018fbi}, is not
as effective. Since this abstraction tends to lose some important features encoded in 
the joints' spatial domain. Instead of elevating the
problem straight away to the 3D volumetric space, we utilize an
intermediate representation of the 3D-aware relationship of the parent joint $p_k$
and the child joint $c_k$ of a skeletal part $B_k$. Provided with the supervision signals, we define
polarized target heatmaps where a pair of normalized Gaussian peeks
corresponding to the 2D joint locations are
placed accordingly across three heatmaps (see Figure~\ref{fig:tri_heatmaps}).
We term them as the \textit{negative polarity heatmap} ${\bf T}^{-1}_k$, the \textit{zero polarity heatmap} ${\bf T}^{0}_k$ and the
\textit{positive polarity heatmap} ${\bf T}^{+1}_k$ with respect to the function value in
Eq.~(\ref{eq:polarity}). The parent joint $p_k$ is always placed in the
zero polarity heatmap ${\bf T}^{0}_k$. The child joint $c_k$ will appear in the negative/positive
polarity heatmap, if its depth is larger/smaller than that of the parent joint $p_k$~(i.e., $|r(z_p,z_c)| \neq 0$).
Both parent and child joints are co-located in the zero polarity heatmap if their
depths are roughly the same~(i.e., $r(z_p,z_c)=0$).

Formally, we denote the heatmap triplets of the skeletal part $B_k$ as the stacking
of three heatmaps ${\bf T}^{-1}_k,{\bf T}^{0}_k,{\bf T}^{+1}_k$: 
\begin{equation}
{\bf T}_k = \operatorname{Stack}[{\bf T}^{-1}_k,{\bf T}^{0}_k,{\bf T}^{+1}_k] .
\end{equation}

Given 3D groundtruth coordinates of all joints, 
we can readily compute the heatmap triplets of each skeletal part. For 
easy reference, we shall refer to the part-centric heatmap triplets 
${\bf T}_k$ as \textit{HEMlets}, and use it afterwards. \\  

\textbf{\textit {Discussions.}}
Here we provide some understandings of HEMlets from a few perspectives. First, different from a joint-specific 2D heatmap that models the detection likelihood for each intended joint on the $(x, y)$ plane, HEMlets models part-centric pairwise joints' co-location likelihoods on the $(x, y)$ plane simultaneously with their ordinal depth relations. This helps to learn geometric constraints (e.g., bone lengths) implicitly. Second, by augmenting a 2D heatmap to a triplet of heatmaps, HEMlets learns and evaluates the co-location likelihood for a pair of connected joints $(p,c)$ by the joint probability distribution $P(x_p,y_p,x_c,y_c,r(z_p,z_c))$ in a locally-defined volumetric space. In contrast, Pavlakos~{\it et al}.~\cite{pavlakos2018ordinal} relaxed the learning target and marginalized the 3D probability distributions independently for the $(x, y)$ plane i.e., $P(x_p,y_p), P(x_c,y_c)$ and the $z$-dimension, with the latter supervised independently by $r(z_p,z_c)$ based on a ranking loss. Third, by exploiting the available supervision signals to a larger extent, HEMlets brings the benefit of making the knowledge more explicitly expressed and easier to learn, and bridges the gap in learning the 3D information from a given 2D image.


\subsection{3D pose inference}
\noindent 
\textbf{\textit {Network architecture.}}
We employ a fully convolutional network to predict the 3D human pose as 
illustrated in Figure~\ref{fig:framework}. A ResNet-50~\cite{ghiasi2014occlusion} 
backbone architecture is adopted for basic feature extraction. 
One of the two upsampling branches is used to learn the HEMlets 
and the 2D heatmaps of skeletal joints, and the other one 
is used to perform upsampling of the learned features to the same resolution 
as the output heatmaps. Both HEMlets and the 2D joint heatmaps are then 
encoded jointly by a 2D convolutional operation to form a latent global 
representation. Finally these global features are joined with the convolutional features 
extracted from the original image to predict a 3D feature map for each joint. 
We perform a soft-argmax operation~\cite{sun2017integral} to aggregate 
information in the 3D feature maps to obtain the 3D joint estimations. \\

\label{sec:training}
\vspace{-3pt}
\Paragraph{\textit {HEMlets loss.}}
Let us denote with ${\bf T}^{\rm gt}$ the groundtruth HEMlets of all skeletal parts 
and with ${\bf \hat{T}}$ the corresponding prediction. We use a standard $L_2$ 
distance between ${\bf T}^{\rm gt}$ and ${\bf \hat{T}}$ to compute the HEMlets loss as follows:
\begin{equation}
{\mathcal{L}}^{\rm HEM} = {\lVert ({\bf T}^{\rm gt} - {\bf \hat{T}}) \odot {\bf \Lambda} \rVert }_2^2 ,
\label{eq:tri-state-loss}
\end{equation}
where $\odot$ denotes an element-wise multiplication, and ${\bf \Lambda}$ is a binary tensor 
to mask out missing annotations.\\

\noindent \textbf{\textit {Auxiliary 2D joint loss.}}
As HEMlets essentially contains heatmap responses of 2D joint locations, we 
adopt a heatmap-based 2D joint detection scheme to facilitate HEMlets prediction. 
The $L_2$ loss of 2D joint prediction is computed as:
\begin{equation}
\mathcal{L}^{\rm 2D} = \sum_{n=1}^N{\lVert {\bf H}^{\rm gt}_n -  {\bf \hat{H}}_n \rVert}_2^2,
\label{eq:heat-loss}
\end{equation}
where ${\bf H}^{\rm gt}_n$ is the groundtruth 2D heatmap of the $n$-th 2D joint 
and ${{\bf{\hat{H}}}_n}$ is the corresponding network prediction.\\

\vspace{-4pt}
\Paragraph{\textit {Soft-argmax 3D joint loss.}}
To avoid quantization errors and allow end-to-end learning, Sun~{\it et al}.~\cite{sun2017integral} 
suggested a soft-argmax regression for 3D human pose estimation. Given learned
volumetric features ${\bf F}_n$ of size $(h \times w \times d)$ for the $n$-th joint,
the predicted 3D coordinates are given as:

\begin{equation}
[\hat{x}_n,\hat{y}_n,\hat{z}_n] = \int_{\mathbf{v}}{ \mathbf{v} \cdot \operatorname{Softmax}({\bf F}_n)},
\label{eq:Integral}
\end{equation}
where $\mathbf{v}$ denotes a voxel in the volumetric feature space of ${\bf F}_n$. For robustness, we employ the $L_1$ loss for the regression of 3D joints. Specifically, the loss is defined as:

\begin{equation}
\mathcal{L}^{\rm 3D}_{\lambda} = \sum_{n=1}^N{ ( \left| x_n^{\rm gt} -  {\hat{x}}_n\right| + \left| y_n^{\rm gt} -  {\hat{y}}_n\right| + \lambda \left| z_n^{\rm gt} -  {\hat{z}}_n\right| )},
\label{eq:3d-loss-lambda}
\end{equation}
where the groundtruth 3D position of the  $n$-th joint is given as $(x_n^{\rm gt},y_n^{\rm gt},z_n^{\rm gt})$. We use the same 2D and 3D mixed training strategy in~\cite{sun2017integral}~($\lambda \in \{0,1\}$): $\lambda$ in Eq.~(\ref{eq:3d-loss-lambda}) is set to $1$ when the training data is from 3D datasets, and $\lambda = 0$ when the data is from 2D datasets. \\

\vspace{-4pt}
\Paragraph{\textit {Training strategy.}}
For HEMlets prediction, We combine $\mathcal{L}^{\rm HEM}$ and $\mathcal{L}^{\rm 2D}$ for the intermediate supervision. 
The loss function is defined as:
\begin{equation}
\mathcal{L}^{\rm int} =  \mathcal{L}^{\rm HEM} + \mathcal{L}^{\rm 2D}.
\label{eq:3d-loss}
\end{equation}
By using $\mathcal{L}^{\rm HEM}$ and $\mathcal{L}^{\rm 2D}$ jointly as supervisions, we allow training the 
network using images with 2D joint annotations and 3D joint annotations. By 
3D joint annotation, we refer to annotations with exact 3D joint coordinates or 
relative depth ordering between part-centric joint pairs. 

The end-to-end training loss $\mathcal{L}^{\rm tot}$ is defined by combining $\mathcal{L}^{\rm int}$ with $\mathcal{L}^{\rm 3D}_{\lambda}$:
\begin{equation}
\mathcal{L}^{\rm tot} = \alpha * \mathcal{L}^{\rm int}  + \mathcal{L}^{\rm 3D}_{\lambda},
\label{eq:mix-loss}
\end{equation}
where $\alpha=0.05$ in all our experiments.\\

\vspace{-4pt}
\Paragraph{\textit{Implementation details.}}
We implement our method in PyTorch. The model is trained in an end-to-end manner 
using both images with 3D annotations (e.g., Human3.6M~\cite{ionescu2014human3} or   
HumanEva-I~\cite{sigal2010humaneva}), and 2D annotations~(MPII~\cite{Andriluka}). 
In our experiments, we adopt an adaptive value of $\epsilon$ in 
Eq.~(\ref{eq:polarity}) for each skeletal part: 
${\epsilon}_k = 0.5 {\lVert B_k \rVert}$~($ {\lVert B_k \rVert}$ is the 
3D Euclidean distance between the two end joints of the skeletal part $B_k$).
The training data is further augmented with rotation ($\pm30^{\circ}$), 
scale ($0.75\!-\!1.25$), horizontal flipping (with a probability of $0.5$) and color distortions.  
By using a batch size of $64$, a learning rate of $0.001$ and Adam optimization, the 
training took $100K$ iterations to converge. It took about a few days~($2\!-\!4$) with 
four NVIDIA GTX 1080 GPUs to train the model. \\



\begin{table*}[ht]

\setlength{\tabcolsep}{2pt}

\small

\resizebox{\textwidth}{50mm}{

\begin{tabular}{lcccccccccccccccc}
\vspace{4pt}
\\ \hline

\textbf{Protocol \#1}          & Direct        & Discuss       & Eating        & Greet         & Phone         & Photo         & Pose          & Purch.      & Sitting       & SittingD.     & Smoke         & Wait          & WalkD.        & Walk          & WalkT.        & \textbf{Avg}   \\ \hline

LinKDE~{\it et al}.~\cite{ionescu2014human3}   &132.7     &183.6    &132.3    &164.4    &162.1    &205.9    &150.6    &171.3    &151.6    &243.0    &162.1    &170.7    &177.1    &96.6    &127.9    &162.1 \\
Tome~{\it et al}..~\cite{tome2017lifting}       &65.0      &73.5     &76.8     &86.4     &86.3     &110.7    &68.9     &74.8     &110.2    &173.9    &85.0     &85.8     &86.3     &71.4    &73.1     &88.4  \\
Rogez~{\it et al}.~\cite{rogez2017lcr}		  &76.2      &80.2     &75.8     &83.3     &92.2     &105.7    &79.0     &71.7     &105.9    &127.1    &88.0     &83.7     &86.6     &64.9    &84.0     &87.7  \\
Tekin~{\it et al}.~\cite{tekin2017learning}    &54.2      &61.4     &60.2     &61.2     &79.4     &78.3     &63.1     &81.6     &70.1     &107.3    &69.3     &70.3     &74.3     &51.8    &74.3     &69.7  \\
Martinez~{\it et al}.~\cite{martinez2017simple}&53.3      &60.8     &62.9     &62.7     &86.4     &82.4     &57.8     &58.7     &81.9     &99.8     &69.1     &63.9     &67.1     &50.9    &54.8     &67.5  \\

Fang~{\it et al}.~\cite{fang2018learning}         &50.1    &54.3    &57.0    &57.1    &66.6    &73.3    &53.4    &55.7    &72.8    &88.6    &60.3    &57.7    &62.7    &47.5    &50.6    &60.4  \\
Pavlakos~{\it et al}.~\cite{pavlakos2018ordinal}  &48.5    &54.4    &54.4    &52.0    &59.4    &65.3    &49.9    &52.9    &65.8    &71.1    &56.6    &52.9    &60.9    &44.7    &47.8    &56.2  \\
S{\'a}r{\'a}ndi~{\it et al}.~\cite{sarandi2018robust}&51.2 &58.7    &51.7    &53.4    &56.8    &59.3    &50.7    &52.6    &65.5    &73.2    &56.8    &51.4    &56.6    &47.0    &42.4    &55.8  \\
Sun~{\it et al}.~\cite{sun2017integral}           &47.5    &47.7    &49.5    &50.2    &51.4    &55.8    &43.8    &46.4    &58.9    &65.7    &49.4    &47.8    &49.0    &38.9    &43.8    &49.6  \\ \hline

Ours   & \textbf{34.4}  & \textbf{42.4}  & \textbf{36.6}  & \textbf{42.1}  & \textbf{38.2}  & \textbf{39.8}  & \textbf{34.7}  & \textbf{40.2}  & \textbf{45.6}  & \textbf{60.8}  & \textbf{39.0}  & \textbf{42.6}  &\textbf{42.0}  & \textbf{29.8}  & \textbf{31.7}  & \textbf{39.9}\\ \hline 

\hline

%
%

\textbf{Protocol \#2}          & Direct        & Discuss       & Eating        & Greet         & Phone         & Photo         & Pose          & Purch.      & Sitting       & SittingD.     & Smoke         & Wait          & WalkD.        & Walk          & WalkT.        & \textbf{Avg}   \\ \hline

Nie~{\it et al}.~\cite{nie2017monocular}          &90.1    &88.2    &85.7    &95.6    &103.9   &92.4    &90.4    &117.9   &136.4   &98.5    &103.0   &94.4    &86.0    &90.6    &89.5    &97.5       \\
Chen~{\it et al}..~\cite{chen20173d}               &53.3    &46.8    &58.6    &61.2    &56.0    &58.1    &41.4    &48.9    &55.6    &73.4    &60.3    &45.0    &76.1    &62.2    &51.1    &57.5       \\
Martinez~{\it et al}.~\cite{martinez2017simple}   &39.5    &43.2    &46.4    &47.0    &51.0    &56.0    &41.4    &40.6    &56.5    &69.4    &49.2    &45.0    &49.5    &38.0    &43.1    &47.7       \\
Fang~{\it et al}.~\cite{fang2018learning}         &38.2    &41.7    &43.7    &44.9    &48.5    &55.3    &40.2    &38.2    &54.5    &64.4    &47.2    &44.3    &47.3    &36.7    &41.7    &45.7  \\
Pavlakos~{\it et al}..~\cite{pavlakos2018ordinal}  &34.7    &39.8    &41.8    &38.6    &42.5    &47.5    &38.0    &36.6    &50.7    &56.8    &42.6    &39.6    &43.9    &32.1    &36.5    &41.8  \\
Yang~{\it et al}.~\cite{yang20183d}               & \textbf{26.9}    &\textbf{30.9}    &36.3    &39.9    &43.9    &47.4    &28.8    &\textbf{29.4}    &\textbf{36.9}    &58.4    &41.5    &\textbf{30.5}    &\textbf{29.5}    &42.5    &32.2    &37.7  \\ \hline
Ours   &29.1  & 34.9  & \textbf{29.9}  & \textbf{32.6}  & \textbf{31.2}  & \textbf{32.3}  & \textbf{27.0}  & 33.3  & 37.6  & \textbf{45.9}  & \textbf{32.2}  &31.5  &34.5  & \textbf{22.9}  & \textbf{25.9}  & \textbf{32.1}\\ \hline

\hline

\textbf{PA MPJPE}          & Direct        & Discuss       & Eating        & Greet         & Phone         & Photo         & Pose          & Purch.      & Sitting       & SittingD.     & Smoke         & Wait          & WalkD.        & Walk          & WalkT.        & \textbf{Avg}   \\ \hline

Yasin~{\it et al}.~\cite{yasin2016dual}  &88.4    &72.5    &108.5    &110.2    &97.1    &81.6    &107.2    &119.0    &170.8    &108.2    &142.5    &86.9    &92.1    &165.7    &102.0    &108.3    \\
Sun~{\it et al}..~\cite{sun2017integral}  &36.9    &36.2    &40.6     &40.4     &41.9    &34.9    &35.7     &50.1     &59.4    &40.4    &44.9    &39.0    &\textbf{30.8}       &39.8     &36.7     &40.6    \\
Dabral~{\it et al}.~\cite{dabral2018learning} &28.0    &30.7    &39.1     &34.4     &37.1    &44.8    &28.9     &32.2     &39.3    &60.6    &39.3    &31.1    &37.8      &25.3     &28.4     &36.3    \\ \hline
Ours   & \textbf{21.6}  & \textbf{27.0}  & \textbf{29.7}  & \textbf{28.3}  & \textbf{27.3}  & \textbf{32.1}  & \textbf{23.5}  & \textbf{30.3}  & \textbf{30.0}  & \textbf{37.7}  & \textbf{30.1}  & \textbf{25.3}  & 34.2 &\textbf{19.2}  & \textbf{23.2}  & \textbf{27.9}\\ \hline
\end{tabular}

}

\vspace{2pt}
\caption{Quantitative comparisons of the mean per-joint position error (MPJPE) on Human3.6M~\cite{ionescu2014human3} under Protocol~\#1 and Protocol~\#2, as well as using PA MPJPE as the evaluation metric. Similar to most of the competing methods (e.g.,~\cite{sun2017integral,pavlakos2018ordinal,yang20183d,dabral2018learning,tekin2017learning,fang2018learning}), our models were trained on the Human3.6M dataset and used also the extra MPII 2D pose dataset~\cite{Andriluka}.} 

\label{table:protocols}
\end{table*}

\Section{Experiments} \label{sec:Experimental-Result}
We perform quantitative evaluation on three benchmark datasets: Human3.6M~\cite{ionescu2014human3}, HumanEva-I~\cite{sigal2010humaneva} and MPI-INF-3DHP~\cite{mehta2017monocular}. 
Ablation study is conducted to evaluate our design choices. 
We demonstrate that the proposed method shows superior generalization ability to 
in-the-wild images. 

\subsection{Datasets and evaluation protocols}

\Paragraph{\textit{Human3.6M.}}
Human3.6M~\cite{ionescu2014human3} contains 3.6 million RGB images captured by a MoCap System in 
an indoor environment, in which 7 professional actors were performing 
15 activities such as walking, eating, sitting, making a phone call and 
engaging in a discussion, etc. 
We follow the standard protocol as in~\cite{martinez2017simple, pavlakos2017coarse}, 
and use 5 subjects (S1, S5, S6, S7, S8) for training and the rest 2 subjects (S9, S11) 
for evaluation (referred to as Protocol~\#1). 
Some previous works reported their results with 6 subjects 
(S1, S5, S6, S7, S8, S9) used for training and only S11 for evaluation~\cite{yasin2016dual,sun2017integral,dabral2018learning} (referred to as Protocol~\#2). Despite {\it not} using S9 as training data, 
we compare our results with these methods.

\Paragraph{\textit{HumanEva-I.}} 
HumanEva-I~\cite{sigal2010humaneva} is one of the early datasets for evaluating 3D human poses. 
It contains fewer subjects and actions compared to Human3.6M. 
Following~\cite{Bo2010Twin}, we train a single model on 
the training sequences of Subject 1, 2 and 3, and evaluate on 
the validation sequences.

\Paragraph{\textit{MPI-INF-3DHP.}} 
This is a recent 3D human pose dataset which includes both indoor and outdoor scenes~\cite{mehta2017monocular}. 
Without using its training set, we evaluate our model trained from Human3.6M only 
on the test set. The results are reported using the 3DPCK and the AUC metric~\cite{Andriluka,mehta2017monocular,pavlakos2018ordinal}.

\Paragraph{\textit{Evaluation metric.}}
We follow the standard steps to align the 3D pose prediction with the groundtruth by 
aligning the position of the central hip joint, and use 
the \textit{Mean Per-Joint Position Error}~(MPJPE) between 
the groundtruth and the prediction as evaluation metrics. 
In some prior works~\cite{yasin2016dual,sun2017integral,dabral2018learning}, 
the pose prediction was further aligned with the 
groundtruth via a rigid transformation. The resulting MPJPE is termed as \textit{Procrustes Aligned}~(PA) MPJPE. 

\subsection{Results and comparisons}
\label{sec:stoa}
\Paragraph{\textit{Human3.6M.}} 
We compare our method against  state-of-the-art under three protocols, and the quantitative 
results are reported in Table~\ref{table:protocols}. As can be seen,
our method outperforms all competing methods on all action subjects 
for the protocols used. It is worth mentioning that our approach 
makes considerable improvements on some challenging actions for 3D pose 
estimation such as {\it Sitting} and {\it Sitting Down}. 
Thanks to HEMlets learning, our method demonstrates a clear advantage for handling 
complicated poses. 

With a simple network architecture and little parameter 
tuning, we produce the most competitive results compared to previous 
works with carefully designed networks powered by e.g., adversarial training schemes 
or prior knowledge. On average, we improve the 3D pose prediction accuracy 
by $20\%$ than that reported in Sun~{\it et al}.~\cite{sun2017integral} under Protocol \#1. 
We also report our performance using PA MPJPE as the evaluation metric, and 
compare with these methods that make use of S9 as additional training data. We 
still outperform all of them across all action subjects, even {\it without} 
utilizing S9 for training. 

\begin{table}[t]
\small
\setlength{\tabcolsep}{2pt}
  \centering
	
  \begin{tabular}{c|ccc|ccc|c}
\hline

\multirow{2}{*}{Approach}&
\multicolumn{3}{c}{Walking}&
\multicolumn{3}{c}{Jogging}\\
									&S1    &S2    &S3    &S1    &S2    &S3    &\textbf{Avg}\\ \hline
\vspace{2pt}
Simo-Serra~{\it et al}.~\cite{simo2013joint}         &65.1  &48.6  &73.5  &74.2  &46.6  &32.2  &56.7 \\ 
Moreno-Noguer~{\it et al}.~\cite{moreno20173d}       &19.7  &13.0  &24.9  &39.7  &20.0  &21.0  &26.9 \\ 
Martinez~{\it et al}.~\cite{martinez2017simple}      &19.7  &17.4  &46.8  &26.9  &18.2  &18.6  &24.6 \\ 
Fang~{\it et al}.~\cite{fang2018learning}            &19.4  &16.8  &37.4  &30.4  &17.6  &16.3  &22.9 \\
Pavlakos~{\it et al}.~\cite{pavlakos2018ordinal}     &18.8  &12.7  &29.2  &\textbf{23.5}  &15.4  &14.5  &18.3 \\ \hline
Ours                                           &\textbf{13.5}  &\textbf{9.9}  &\textbf{17.1}   &24.5 &\textbf{14.8}  &\textbf{14.4}  &\textbf{15.2} \\ \hline

\end{tabular}
\vspace{4pt}
\caption{Detailed results on the validation set of HumanEva-I~\cite{mehta2017monocular}. }
\label{table:HumanEva-I}
\end{table}

\Paragraph{\textit{HumanEva-I.}} 
With the same network architecture where {\it only} the HumanEva-I dataset is used for training, our results are reported in Table~\ref{table:HumanEva-I} under the popular protocol~\cite{simo2013joint,moreno20173d,martinez2017simple,fang2018learning,pavlakos2018ordinal}. Different from these approaches~\cite{pavlakos2018ordinal,moreno20173d,martinez2017simple,fang2018learning} which used extra 2D datasets~(e.g., MPII) or pre-trained 2D detectors~(e.g., CPM~\cite{wei2016convolutional}), our method still outperforms previous approaches.

\begin{table}[t]
\small
\setlength{\tabcolsep}{2pt}
  \centering
	
  \begin{tabular}{cccccc}
\hline
\vspace{4pt}

\multirow{3}{*}{Approach}

&Studio  & Studio &

\multirow{2}{*}{Outdoor  }&
\multirow{2}{*}{All  }&
\multirow{2}{*}{All  } \\

&GS &no GS \\ 

\cline{2-6}
\vspace{2pt}
&3DPCK   &3DPCK   &3DPCK   &3DPCK  &AUC\\

\hline

Mehta~{\it et al}.~\cite{mehta2017monocular}     &70.8    &62.3    &58.8    &64.7    &31.7\\
Zhou~{\it et al}.~\cite{zhou2017towards}         &71.1    &64.7    &72.7    &69.2    &32.5\\
Pavlakos~{\it et al}.~\cite{pavlakos2018ordinal} &\textbf{76.5}    &63.1    &77.5    &71.9    &35.3\\
\hline
ours                                       &75.6    &\textbf{71.3}    &\textbf{80.3}    &\textbf{75.3}    &\textbf{38.0}\\
\hline
\end{tabular}
\vspace{4pt}
\caption{Detailed results on the test set of MPI-INF-3DHP~\cite{mehta2017monocular}. No training data from this dataset was used to train our model.}
\label{table:3dhpResults}
\end{table}

\Paragraph{\textit{MPI-INF-3DHP.}}
We evaluate our method on the MPI-INF-3DHP dataset using two metrics, the PCK and AUC. The results are generated by the model we trained for Human3.6M. In Table~\ref{table:3dhpResults}, we compare with three recent methods which are not trained on this dataset. Our result of ``Studio GS" is one percentage lower than~\cite{pavlakos2018ordinal}. But our method outperforms all these methods with particularly large margins for the ``Outdoor" and ``Studio no GS" sequences.

\subsection{Ablation study}
\label{sec:ablation}

We study the influence on the final estimation performance of different choices made in our network design and the training procedure.

\vspace{-2pt}
\Paragraph{\textit{Alternative intermediate supervision.}}  
First, We examine the effectiveness of using HEMlets supervision. 
We evaluate the model trained without any intermediate supervision (Baseline), 
with 2D heatmap supervision only, with HEMlets supervision only, and 
with both 2D heatmap supervision and HEMlets supervision (Full). 
All of these design variants are evaluated with the same experimental 
setting (including training data, network architecture and $\mathcal{L}^{\rm 3D}_{\lambda}$ loss 
definition) under Protocol \#1 on Human3.6M. 


\begin{table}[t]
\small
\renewcommand\tabcolsep{3.0pt} 
\begin{center}
\begin{tabular}{llcc}
\hline
Method & Supervision & H3.6M \#1  & H3.6M \#1$^*$ \\

\hline

 Baseline               &  $\mathcal{L}^{\rm 3D}_{\lambda}$& 47.1  &55.3\\
w/ 2D heatmaps                    & $\mathcal{L}^{\rm 3D}_{\lambda} + \mathcal{L}^{\rm 2D}$ & 44.2  & 49.9\\
w/   HEMlets                & $\mathcal{L}^{\rm 3D}_{\lambda} + \mathcal{L}^{\rm HEM}$& 42.6		& 46.0\\
   Full                & $\mathcal{L}^{\rm 3D}_{\lambda} + \mathcal{L}^{\rm HEM} + \mathcal{L}^{\rm 2D}$ & 39.9 & 45.1\\

\hline
\end{tabular}
\end{center}
\caption{Ablative study on the effects of alternative intermediate supervision evaluated on Human3.6M using Protocol~\#1. The last column~$^*$ reports the results using only the Human3.6M dataset for training (without using the extra MPII 2D pose dataset).}
\label{table:ablationStudy}
\end{table}

The detailed results are presented in Table~\ref{table:ablationStudy}. Using 2D heatmaps supervision for training, the prediction error is reduced by 3.0mm compared to the baseline. The HEMlets supervision provided 1.7mm lower mean error compared to the 2D heatmaps supervision. This validates the effectiveness of the intermediate supervision. By combining all these choices, our approach using HEMlets with 2D heatmap supervision achieves the lowest error. Without using the extra MPII 2D pose dataset, we repeated this study. Similar conclusions can still be drawn. But the gap between w/ HEMlets (excluding $\mathcal{L}^{\rm 2D}$, 46.0mm) and Full (45.1mm) shrinks, suggesting the strength of the HEMlets representation in encoding both 2D and (local) 3D information.

\begin{figure}[t]
\centering
\vspace{-0.1in}
\includegraphics[width = \columnwidth ]{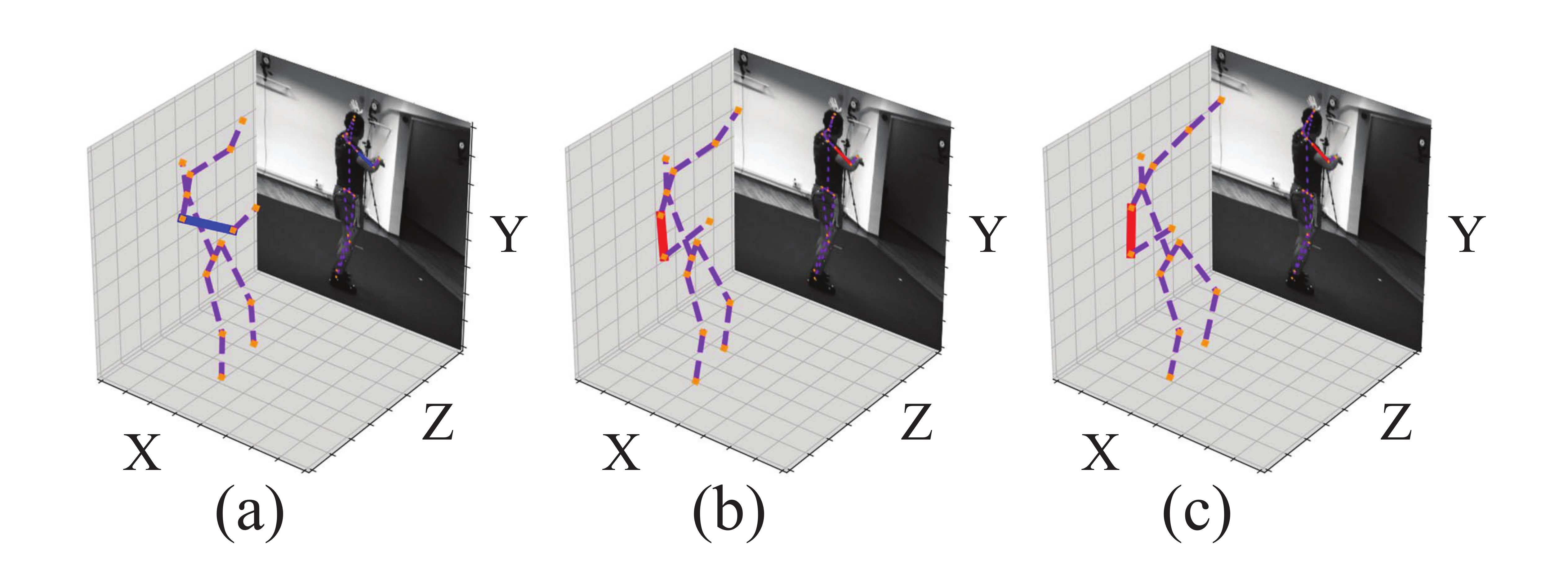}
\caption{An example image with the detected joints overlaid and shown from a novel view, using different methods: (a) $\mathcal{L}^{\rm 3D}_{\lambda} + \mathcal{L}^{\rm 2D}$ (2D error:~15.2; 3D joint error:~{\bf81.3mm}). (b) $\mathcal{L}^{\rm 3D}_{\lambda} + \mathcal{L}^{\rm 2D} + \mathcal{L}^{\rm HEM}$ (2D error:~13.0; 3D error:~{\bf41.2mm}). (c) Ground-truth. HEMlets learning helps fixing local part errors, see {\color{blue} blue} in~(a) vs. {\color{red} red} in~(b).}
\label{fig:example}
\end{figure}

To further illustrate the effectiveness of HEMlets representation, we provide a visual comparison in Fig.~\ref{fig:example}. Though the 2D joint errors of the two estimations are quite close, the method with HEMlets learning significantly improves the 3D joint estimation result and fixes the gross limb errors.

Regarding the runtime, tested on a NVIDIA GTX 1080 GPU, our full model (with a total parameter number of 47.7M) takes 13.3ms for a single forward inference, while the baseline model (with 34.3M parameters) takes 8.5ms.

\begin{figure*}[t]
  \centering
  \includegraphics[width=0.96\linewidth]{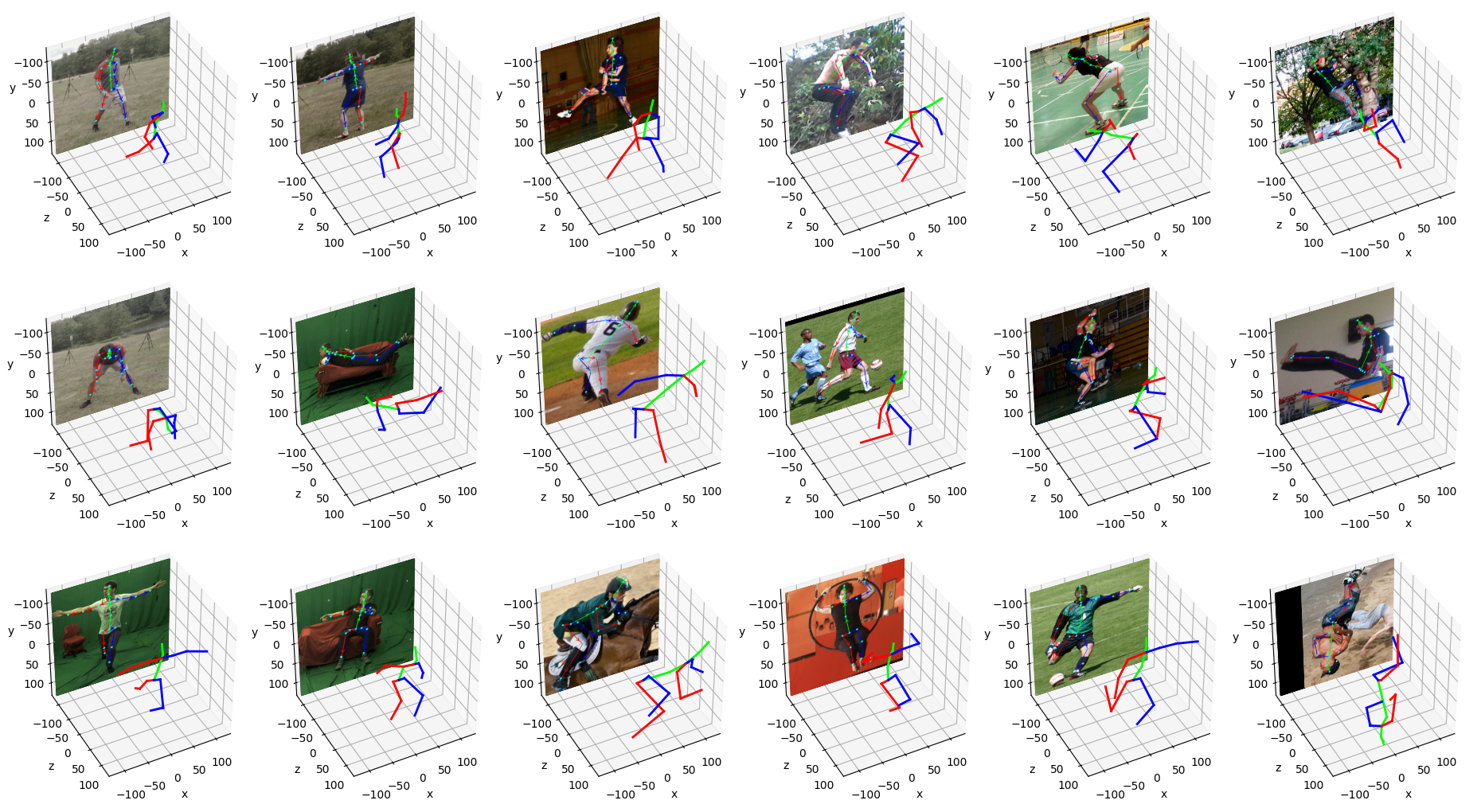}
  \caption{Qualitative results on different validation datasets: the first two columns are from the test dataset of 3DHP~\cite{mehta2017monocular}. The other columns are from Leeds Sports Pose~(LSP)~\cite{johnson2010clustered}. Our approach produces visually correct results even on challenging poses~(last column). }
\label{fig:wild}
\end{figure*}

\begin{figure}[h]
\vspace*{-7pt}
  \centering
	\vspace{-2pt}
  \includegraphics[width=0.80\columnwidth]{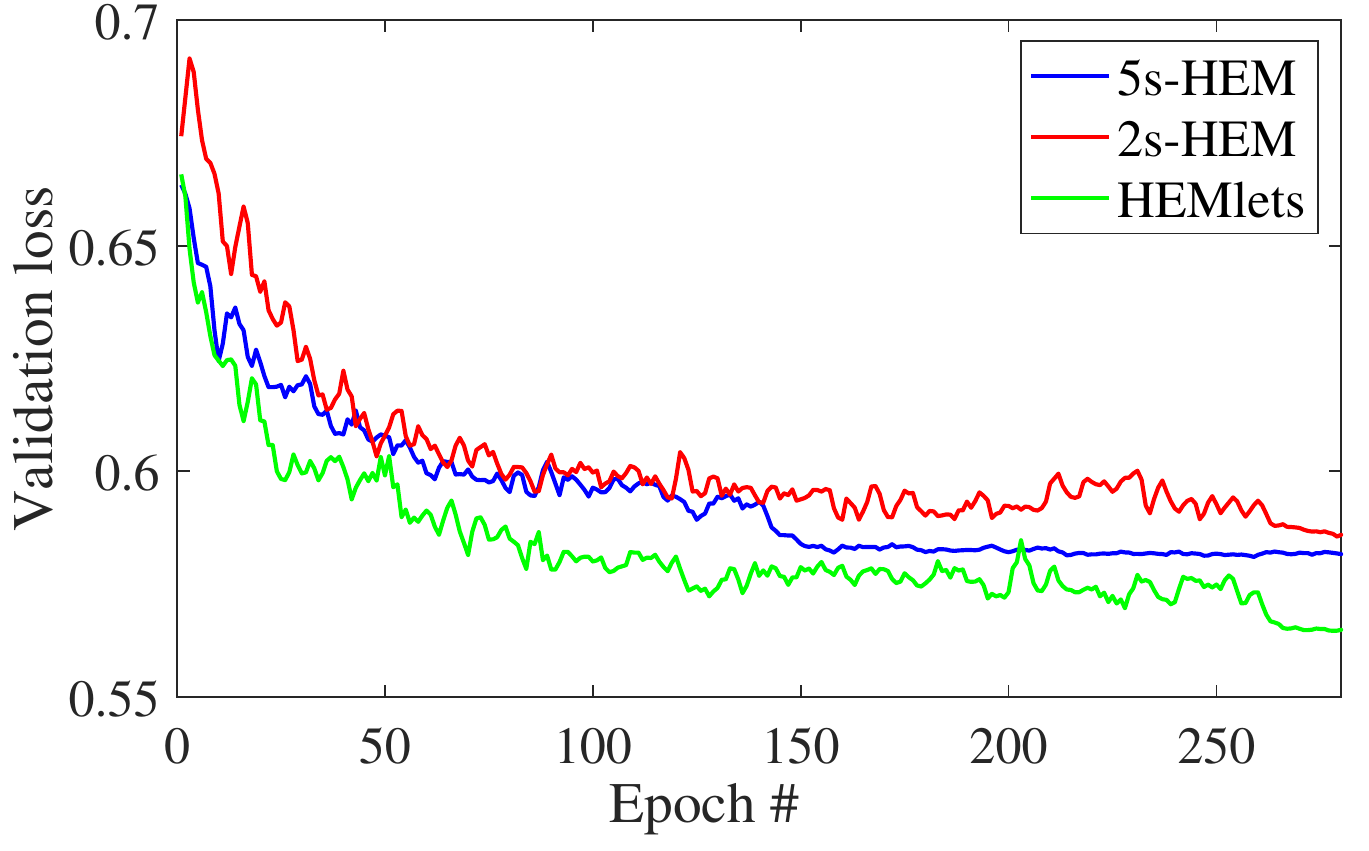} 
  \caption{The validation loss of 5s-HEM, 2s-HEM and HEMlets, respectively. All are trained with the Human3.6M dataset. } 
\label{fig:Fiv-tri-state result}
\end{figure}

\Paragraph{\textit{Variants of HEMlets.}}
We next experimented with some variants of HEMlets on Human3.6M and MPII 2D pose datasets. In the first variant, we use five-state heatmaps, 
referred to as \textit{5s-HEM}, where the child joint is placed to different layers of the 
heatmaps according to the angle of the associated skeletal part with respect to the imaging plane. 
Specifically, we 
define the five states corresponding to the $(-90^\circ,-60^\circ)$, $(-60^\circ, -30^\circ)$, 
$(-30^\circ, 30^\circ)$, $(30^\circ, 60^\circ)$ and $(60^\circ, 90^\circ)$ range, respectively. 
In the second variant, we place a pair of joints in the negative and positive polarity
heatmaps respectively according to their depth ordering (i.e., the closer/farther joint will appear in the positive/negative polarity heatmap. If their depths are roughly the same, they are co-located in the zero polarity heatmap. We refer to this variant as \textit{2s-HEM}.  We trained 5s-HEM, 2s-HEM and HEMlets with the Human3.6M dataset only. A comparison on the validation loss is given in Fig.~\ref{fig:Fiv-tri-state result}. 
The other two variants produce inferior convergence compared to HEMlets under the same experiment setting. 

\vspace{-2pt}
\Paragraph{\textit{Augmenting datasets.}}
Many state-of-the-art approaches use a mixed training strategy for 3D human pose estimation. 
In addition to exploiting Human3.6M and MPII datasets, we study the effect of using 
augmenting datasets such as Ordinal~\cite{pavlakos2018ordinal} and FBI~\cite{shi2018fbi} 
for training. Firstly, we adapt the annotations of Ordinal and FBI datasets to the required form of HEMlets. 
Then we train our model using different combinations of these additional datasets. 
The comparisons on the MPI-INF-3DHP dataset~\cite{mehta2017monocular} are reported in Table~\ref{table:DataSetEffectiveness}. We find augmenting datasets 
slightly increase the 3DPCK score for the trained model. Interestingly, training with FBI annotations 
attains a better 3DPCK score than Ordinal annotations. We suspect this is 
due to the amount of manual annotation errors related to different annotation schemes. 

\vspace{-4pt}
\Paragraph{\textit{Generalization.}}
For an evaluation of in-the-wild images from Leeds Sports Pose~(LSP)~\cite{johnson2010clustered} and the validation set of MPI-INF-3DHP~\cite{mehta2017monocular}, we list some visual results predicted by our approach. As shown in Fig.~\ref{fig:wild}, even for challenging data~(e.g., self-occlusion, upside-down), our method yields visually correct pose estimations for these images.  

\Section{Conclusion}
In this paper, we proposed a simple and highly effective HEMlets-based 3D pose estimation method from a single color image. HEMlets is an easy-to-learn intermediate representation encoding the relative forward-or-backward depth relation for each skeletal part's joints, together with their spatial co-location likelihoods. It is proved very helpful to bridge the input 2D image and the output 3D pose in the learning procedure. We demonstrated the effectiveness of the proposed method tested over the standard benchmarks, yielding a relative accuracy improvement of about $20\%$ over the best-of-grade method on the Human3.6M benchmark. Good generalization ability is also witnessed for the presented approach. We believe the proposed HEMlets idea is actually general, which may potentially benefit other 3D regression problems e.g., scene depth estimation.

\begin{table}
\vspace*{-7pt}
\begin{center}
\begin{tabular}{lcc}
\hline
Dataset     &3DPCK\\
\hline
Base                     & 75.3    \\ 
w/  Ordinal~\cite{pavlakos2018ordinal}             & 76.1    \\
w/  FBI~\cite{shi2018fbi}       	    & \textbf{76.9}	   \\

w/  FBI~\cite{shi2018fbi} + Ordinal~\cite{pavlakos2018ordinal}     & 76.5    \\
\hline
\end{tabular}
\end{center}
\caption{Evaluation of 3DPCK scores by adding different augmenting datasets that provide relative depth ordering annotations. Base denotes using the base datasets~(Human3.6M and MPII). }
\label{table:DataSetEffectiveness}
\end{table}


\Paragraph{\bf{Acknowledgements.}}
\small{This work is supported in part by the National Natural Science Foundation of China~(Grant No.:~61771201), the Program for Guangdong Introducing Innovative and Enterpreneurial Teams~(Grant No.:~2017ZT07X183),  the Pearl River Talent Recruitment Program Innovative and Entrepreneurial Teams in 2017~(Grant No.:~2017ZT07X152),  the Shenzhen Fundamental Research Fund~(Grants No.:~KQTD2015033114415450 and ZDSYS201707251409055), and Department of Science and Technology of Guangdong Province Fund~(2018B030338001).}


{\small
\bibliographystyle{ieee_fullname}
\bibliography{HEMletsPose_ICCV19}
}

\end{document}